# VITON: An Image-based Virtual Try-on Network

Xintong Han, Zuxuan Wu, Zhe Wu, Ruichi Yu, Larry S. Davis
University of Maryland, College Park
{xintong,zxwu,zhewu,richyu,lsd}@umiacs.umd.edu

## Abstract

*We present an image-based VIrtual Try-On Network (VITON) without using 3D information in any form, which seamlessly transfers a desired clothing item onto the corresponding region of a person using a coarse-to-fine strategy. Conditioned upon a new clothing-agnostic yet descriptive person representation, our framework first generates a coarse synthesized image with the target clothing item overlaid on that same person in the same pose. We further enhance the initial blurry clothing area with a refinement network. The network is trained to learn how much detail to utilize from the target clothing item, and where to apply to the person in order to synthesize a photo-realistic image in which the target item deforms naturally with clear visual patterns. Experiments on our newly collected dataset demonstrate its promise in the image-based virtual try-on task over state-of-the-art generative models.*[1]

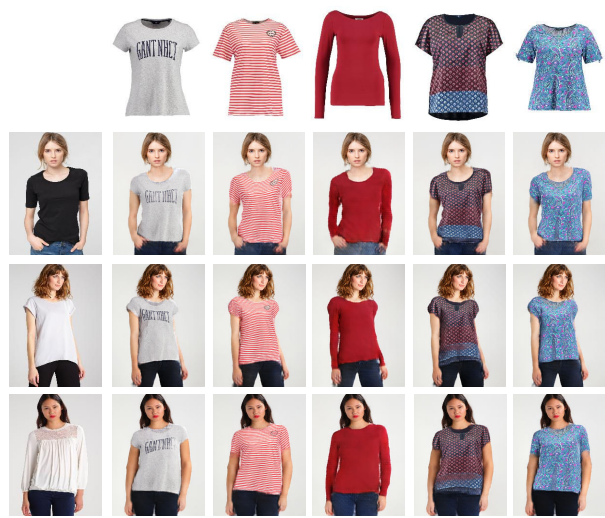

Figure 1: **Virtual try-on results generated by our method**. Each row shows a person virtually trying on different clothing items. Our model naturally renders the items onto a person while retaining her pose and preserving detailed characteristics of the target clothing items.

## 1. Introduction

Recent years have witnessed the increasing demands of online shopping for fashion items. Online apparel and accessories sales in US are expected to reach 123 billion in 2022 from 72 billion in 2016 [1]. Despite the convenience online fashion shopping provides, consumers are concerned about how a particular fashion item in a product image would look on them when buying apparel online. Thus, allowing consumers to virtually try on clothes will not only enhance their shopping experience, transforming the way people shop for clothes, but also save cost for retailers. Motivated by this, various virtual fitting rooms/mirrors have been developed by different companies such as TriMirror, Fits Me, *etc*. However, the key enabling factor behind them is the use of 3D measurements of body shape, either captured directly by depth cameras [40] or inferred from a 2D image using training data [4, 45]. While these 3D modeling techniques enable realistic clothing simulations on the person, the high costs of installing hardwares and collecting 3D annotated data inhibit their large-scale deployment.

We present an image-based virtual try-on approach, relying merely on plain RGB images *without leveraging any 3D information*. Our goal is to synthesize a photo-realistic new image by overlaying a product image seamlessly onto the corresponding region of a clothed person (as shown in Figure 1). The synthetic image is expected to be perceptually convincing, meeting the following desiderata: (1) body parts and pose of the person are the same as in the original image; (2) the clothing item in the product image deforms naturally, conditioned on the pose and body shape of the person; (3) detailed visual patterns of the desired product are clearly visible, which include not only low-level features like color and texture but also complicated graphics like embroidery, logo, *etc*. The non-rigid nature of clothes, which are frequently subject to deformations and occlusions, poses a significant challenge to satisfying these requirements simultaneously, especially without 3D information.

---

[1]Models and code are available at https://github.com/xthan/VITON.

Conditional Generative Adversarial Networks (GANs), which have demonstrated impressive results on image generation [37, 26], image-to-image translation [20] and editing tasks [49], seem to be a natural approach for addressing this problem. In particular, they minimize an adversarial loss so that samples generated from a generator are indistinguishable from real ones as determined by a discriminator, conditioned on an input signal [37, 33, 20, 32]. However, they can only transform information like object classes and attributes roughly, but are unable to generate graphic details and accommodate geometric changes [50]. This limits their ability in tasks like virtual try-on, where visual details and realistic deformations of the target clothing item are required in generated samples.

To address these limitations, we propose a virtual try-on network (VITON), a coarse-to-fine framework that seamlessly transfers a target clothing item in a product image to the corresponding region of a clothed person in a 2D image. Figure 2 gives an overview of VITON. In particular, we first introduce a clothing-agnostic representation consisting of a comprehensive set of features to describe different characteristics of a person. Conditioned on this representation, we employ a multi-task encoder-decoder network to generate a coarse synthetic clothed person in the same pose wearing the target clothing item, and a corresponding clothing region mask. The mask is then used as a guidance to warp the target clothing item to account for deformations. Furthermore, we utilize a refinement network which is trained to learn how to composite the warped clothing item to the coarse image so that the desired item is transfered with natural deformations and detailed visual patterns. To validate our approach, we conduct a user study on our newly collected dataset and the results demonstrate that VITON generates more realistic and appealing virtual try-on results outperforming state-of-the-art methods.

## 2. Related Work

**Fashion analysis**. Extensive studies have been conducted on fashion analysis due to its huge profit potentials. Most existing methods focus on clothing parsing [44, 28], clothing recognition by attributes [31], matching clothing seen on the street to online products [30, 14], fashion recommendation [19], visual compatibility learning [43, 16], and fashion trend prediction [2]. Compared to these lines of work, we focus on virtual try-on with only 2D images as input. Our task is also more challenging compared to recent work on interactive search that simply modifies attributes (*e.g.*, color and textures) of a clothing item [25, 48, 15], since virtual try-on requires preserving the details of a target clothing image as much as possible, including exactly the same style, embroidery, logo, text, *etc*.

**Image synthesis**. GANs [12] are one of most popular deep generative models for image synthesis, and have demonstrated promising results in tasks like image generation [8, 36] and image editing [49, 34]. To incorporate desired properties in generated samples, researchers also utilize different signals, in the form of class labels [33], text [37], attributes [41], *etc*., as priors to condition the image generation process. There are a few recent studies investigating the problem of image-to-image translation using conditional GANs [20], which transform a given input image to another one with a different representation. For example, producing an RGB image from its corresponding edge map, semantic label map, *etc*., or *vice versa*. Recently, Chen and Kolton [6] trained a CNN using a regression loss as an alternative to GANs for this task without adversarial training. These methods are able to produce photo-realistic images, but have limited success when geometric changes occur [50]. Instead, we propose a refinement network that pays attention to clothing regions and deals with clothing deformations for virtual try-on.

In the context of image synthesis for fashion applications, Yoo *et al*. [46] generated a clothed person conditioned on a product image and *vice versa* regardless of the person's pose. Lassner *et al*. [26] described a generative model of people in clothing, but it is not clear how to control the fashion items in the generated results. A more related work is FashionGAN [51], which replaced a fashion item on a person with a new one specified by text descriptions. In contrast, we are interested in the precise replacement of the clothing item in a reference image with a target item, and address this problem with a novel coarse-to-fine framework.

**Virtual try-on**. There is a large body of work on virtual try-on, mostly conducted in computer graphics. Guan *et al*. proposed DRAPE [13] to simulate 2D clothing designs on 3D bodies in different shapes and poses. Hilsmann and P. Eisert [18] retextured the garment dynamically based on a motion model for real-time visualization in a virtual mirror environment. Sekine *et al*. [40] introduced a virtual fitting system that adjusts 2D clothing images to users through inferring their body shapes with depth images. Recently, Pons-Moll *et al*. [35] utilized a multi-part 3D model of clothed bodies for clothing capture and retargeting. Yang *et al*. [45] recovered a 3D mesh of the garment from a single view 2D image, which is further re-targeted to other human bodies. In contrast to relying on 3D measurements to perform precise clothes simulation, in our work, we focus on synthesizing a perceptually correct photo-realistic image directly from 2D images, which is more computationally efficient. In computer vision, limited work has explored the task of virtual try-on. Recently, Jetchev and Bergmann [21] proposed a conditional analogy GAN to swap fashion articles. However, during testing, they require the product images of both the target item and the original item on the person, which makes it infeasible in practical scenar-ios. Moreover, without injecting any person representation or explicitly considering deformations, it fails to generate photo-realistic virtual try-on results.

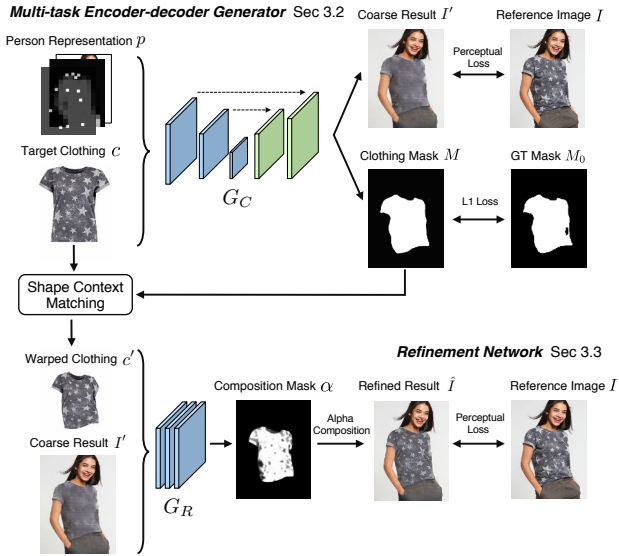

Figure 2: **An overview of VITON**. VITON consists of two stages: (a) an encoder-decoder generator stage (Sec 3.2), and (b) a refinement stage (Sec 3.3).

## 3. VITON

The goal of VITON is, given a reference image $I$ with a clothed person and a target clothing item $c$, to synthesize a new image $\hat{I}$, where $c$ is transferred naturally onto the corresponding region of the same person whose body parts and pose information are preserved. Key to a high-quality synthesis is to learn a proper transformation from product images to clothes on the body. A straightforward approach is to leverage training data of a person with fixed pose wearing different clothes and the corresponding product images, which, however, is usually difficult to acquire.

In a practical virtual try-on scenario, only a reference image and a desired product image are available at test time. Therefore, we adopt the same setting for training, where a reference image $I$ with a person wearing $c$ and the product image of $c$ are given as inputs (we will use $c$ to refer to the product image of $c$ in the following paper). Now the problem becomes given the product image $c$ and the person's information, how to learn a generator that not only produces $I$ during training, but more importantly is able to generalize at test time – synthesizing a perceptually convincing image with an arbitrary desired clothing item.

To this end, we first introduce a clothing-agnostic person representation (Sec 3.1). We then synthesize the reference image with an encoder-decoder architecture (Sec 3.2), conditioned on the person representation as well as the target clothing image. The resulting coarse result is further improved to account for detailed visual patterns and deformations with a refinement network (Sec 3.3). The overall framework is illustrated in Figure 2.

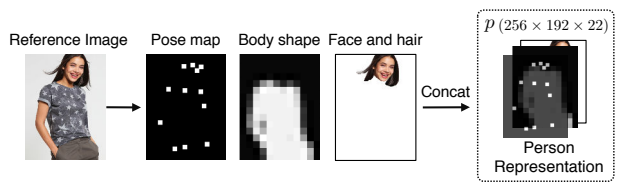

Figure 3: **A clothing-agnostic person representation**. Given a reference image $I$, we extract the pose, body shape and face and hair regions of the person, and use this information as part of input to our generator.

### 3.1. Person Representation

A main technical challenge of a virtual try-on synthesis is to deform the target clothing image to fit the pose of a person. To this end, we introduce a clothing-agnostic person representation, which contains a set of features (Figure 3), including pose, body parts, face and hair, as a prior to constrain the synthesis process.

**Pose heatmap**. Variations in human poses lead to different deformations of clothing, and hence we explicitly model pose information with a state-of-the-art pose estimator [5]. The computed pose of a person is represented as coordinates of 18 keypoints. To leverage their spatial layout, each keypoint is further transformed to a heatmap, with an $11 \times 11$ neighborhood around the keypoint filled in with ones and zeros elsewhere. The heatmaps from all keypoints are further stacked into an 18-channel pose heatmap.

**Human body representation**. The appearance of clothing highly depends on body shapes, and thus how to transfer the target fashion item depends on the location of different body parts (*e.g.*, arms or torso) and the body shape. A state-of-the-art human parser [11] is thus used to compute a human segmentation map, where different regions represent different parts of human body like arms, legs, *etc*. We further convert the segmentation map to a 1-channel binary mask, where ones indicate human body (except for face and hair) and zeros elsewhere. This binary mask derived directly from $I$ is downsampled to a lower resolution ($16 \times 12$ as shown in Figure 3) to avoid the artifacts when the body shape and target clothing conflict as in [51].

**Face and hair segment**. To maintain the identity of the person, we incorporate physical attributes like face, skin color, hair style, *etc*. We use the human parser [11] to extract the RGB channels of face and hair regions of the person to inject identity information when generating new images.

Finally, we resize these three feature maps to the same resolution and then concatenate them to form a clothing-agnostic person representation $p$ such that $p \in \mathrm{R}^{m \times n \times k}$, where $m = 256$ and $n = 192$ denote the height and width of the feature map, and $k = 18 + 1 + 3 = 22$ represents the number of channels. The representation contains abundant information about the person upon which convolutions are performed to model their relations. Note that our representation is more detailed than previous work [32, 51].

### 3.2. Multi-task Encoder-Decoder Generator

Given the clothing-agnostic person representation $p$ and the target clothing image $c$, we propose to synthesize the reference image $I$ through reconstruction such that a natural transfer from $c$ to the corresponding region of $p$ can be learned. In particular, we utilize a multi-task encoder-decoder framework that generates a clothed person image along with a clothing mask of the person as well. In addition to guiding the network to focus on the clothing region, the predicted clothing mask will be further utilized to refine the generated result, as will be discussed in Sec 3.3. The encoder-decoder is a general type of U-net architecture [38] with skip connections to directly share information between layers through bypassing connections.

Formally, let $G_C$ denote the function approximated by the encoder-decoder generator. It takes the concatenated $c$ and $p$ as its input and generates a 4-channel output $(I', M) = G_C(c, p)$, where the first 3 channels represent a synthesized image $I'$ and the last channel $M$ represents a segmentation mask of the clothing region as shown at the top of Figure 2. We wish to learn a generator such that $I'$ is close to the reference image $I$ and $M$ is close to $M_0$ ($M_0$ is the pseudo ground truth clothing mask predicted by the human parser on $I$). A simple way to achieve this is to train the network with an $L_1$ loss, which generates decent results when the target is a binary mask like $M_0$. However, when the desired output is a colored image, $L_1$ loss tends to produce blurry images [20]. Following [22, 27, 7], we utilize a perceptual loss that models the distance between the corresponding feature maps of the synthesized image and the ground truth image, computed by a visual perception network. The loss function of the encoder-decoder can now be written as the sum of a perceptual loss and an $L_1$ loss:

$$L_{G_C} = \sum_{i=0}^{5} \lambda_i ||\phi_i(I') - \phi_i(I)||_1 + ||M - M_0||_1, \quad (1)$$

where $\phi_i(y)$ in the first term is the feature map of image $y$ of the $i$-th layer in the visual perception network $\phi$, which is a VGG19 [42] network pre-trained on ImageNet. For layers $i \geqslant 1$, we utilize 'conv1_2', 'conv2_2', 'conv3_2', 'conv4_2', 'conv5_2' of the VGG model while for layer 0, we directly use RGB pixel values. The hyperparameter $\lambda_i$ controls the contribution of the $i$-th layer to the total loss. The perceptual loss forces the synthesized image to match

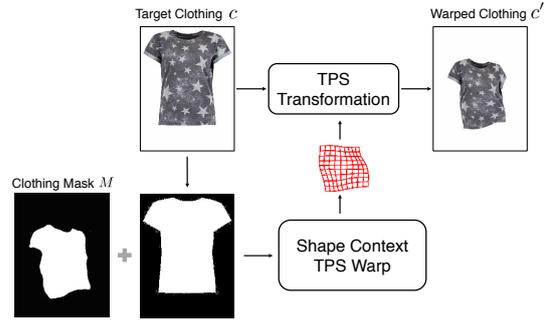

Figure 4: **Warping a clothing image**. Given the target clothing image and a clothing mask predicted in the first stage, we use shape context matching to estimate the TPS transformation and generate a warped clothing image.

RGB values of the ground truth image and their activations at different layers in a visual perception model as well, allowing the synthesis network to learn realistic patterns. The second term in Eqn. 1 is a regression loss that encourages the predicted clothing mask $M$ to be the same as $M_0$.

By minimizing Eqn. 1, the encoder-decoder learns how to transfer the target clothing conditioned on the person representation. While the synthetic clothed person conforms to the pose, body parts and identity in the original image (as illustrated in the third column of Figure 5), details of the target item such as text, logo, *etc*. are missing. This might be attributed to the limited ability to control the process of synthesis in current state-of-the-art generators. They are typically optimized to synthesize images that look similar globally to the ground truth images without knowing where and how to generate details. To address this issue, VITON uses a refinement network together with the predicted clothing mask $M$ to improve the coarse result $I'$.

### 3.3. Refinement Network

The refinement network $G_R$ in VITON is trained to render the coarse blurry region leveraging realistic details from a deformed target item.

**Warped clothing item**. We borrow information directly from the target clothing image $c$ to fill in the details in the generated region of the coarse sample. However, directly pasting the product image is not suitable as clothes deform conditioned on the person pose and body shape. Therefore, we warp the clothing item by estimating a thin plate spline (TPS) transformation with shape context matching [3], as illustrated in Figure 4. More specifically, we extract the foreground mask of $c$ and compute shape context TPS warps [3] between this mask and the clothing mask $M$ of the person, estimated with Eqn. 1. These computed TPS parameters are further applied to transform the target clothing image $c$ into a warped version $c'$. As a result, the warped clothing image conforms to pose and body shape information of the person and fully preserves the details of the target item. The idea is similar to recent 2D/3D texture warping methods for face synthesis [52, 17], where 2D facial keypoints and 3D pose estimation are utilized for warping. In contrast, we rely on the shape context-based warping due to the lack of accurate annotations for clothing items. Note that a potential alternative to estimating TPS with shape context matching is to learn TPS parameters through a Siamese network as in [23]. However, this is particularly challenging for non-rigid clothes, and we empirically found that directly using context shape matching offers better warping results for virtual try-on.

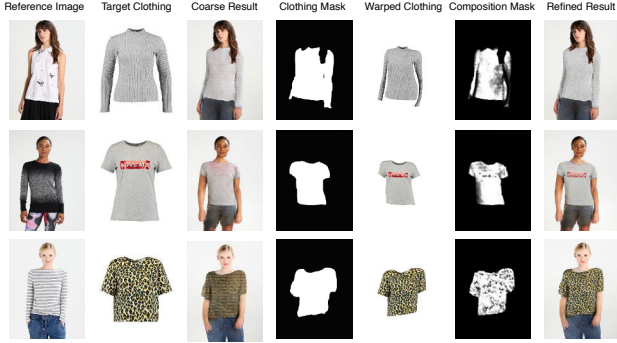

Figure 5: **Output of different steps in our method**. Coarse sythetic results generated by the encoder-decoder are further improved by learning a composition mask to account for details and deformations.

**Learn to composite**. The composition of the warped clothing item $c'$ onto the coarse synthesized image $I'$ is expected to combine $c'$ seamlessly with the clothing region and handle occlusion properly in cases where arms or hair are in front of the body. Therefore, we learn how to composite with a refinement network. As shown at the bottom of Figure 2, we first concatenate $c'$ and the coarse output $I'$ as the input of our refinement network $G_R$. The refinement network then generates a 1-channel composition mask $\alpha \in (0,1)^{m \times n}$, indicating how much information is utilized from each of the two sources, *i.e.*, the warped clothing item $c'$ and the coarse image $I'$. The final virtual try-on output of VITON $\hat{I}$ is a composition of $c'$ and $I'$:

$$\hat{I} = \alpha \odot c' + (1-\alpha) \odot I', \quad (2)$$

where $\odot$ represents element-wise matrix multiplication. To learn the optimal composition mask, we minimize the discrepancy between the generated result $\hat{I}$ and the reference image $I$ with a similar perceptual loss $L_{perc}$ as Eqn. 1:

$$L_{perc}(\hat{I}, I) = \sum_{i=3}^{5} \lambda_i ||\phi_i(\hat{I}) - \phi_i(I)||_1, \quad (3)$$

where $\phi$ denotes the visual perception network VGG19. Here we only use 'conv3_2', 'conv4_2', 'conv5_2' for calculating this loss. Since lower layers of a visual perception network care more about the detailed pixel-level information of an image instead of its content [10], small displacements between $I$ and $\hat{I}$ (usually caused by imperfect warping) will lead to a large mismatch between the feature maps of lower layers ('conv1' and 'conv2'), which, however, is acceptable in a virtual try-on setting. Hence, by only using higher layers, we encourage the model to ignore the effects of imperfect warping, and hence it is able to select the warped target clothing image and preserve more details.

We further regularize the generated composition mask output by $G_R$ with an $L_1$ norm and a total-variation (TV) norm. The full objective function for the refinement network then becomes:

$$L_{G_R} = L_{perc}(\hat{I}, I) - \lambda_{warp}||\alpha||_1 + \lambda_{TV}||\nabla \alpha||_1, \quad (4)$$

where $\lambda_{warp}$ and $\lambda_{TV}$ denote the weights for the $L_1$ norm and the TV norm, respectively. Minimizing the negative $L_1$ term encourages our model to utilize more information from the warped clothing image and render more details. The total-variation regularizer $||\nabla \alpha||_1$ penalizes the gradients of the generated composition mask $\alpha$ to make it spatially smooth, so that the transition from the warped region to the coarse result looks more natural.

Figure 5 visualizes the results generated at different steps from our method. Given the target clothing item and the representation of a person, the encoder-decoder produces a coarse result with pose, body shape and face of the person preserved, while details like graphics and textures on the target clothing item are missing. Based on the clothing mask, our refinement stage warps the target clothing image and predicts a composition mask to determine which regions should be replaced in the coarse synthesized image. Consequentially, important details (material in the 1st example, text in 2nd example, and patterns in the 3rd example) "*copied*" from the target clothing image are "*pasted*" to the corresponding clothing region of the person.

## 4. Experiments

### 4.1. Dataset

The dataset used in [21] is a good choice for conducting experiments for virtual try-on, but it is not publicly available. We therefore collected a similar dataset. We first crawled around 19,000 frontal-view woman and top[2] clothing image pairs and then removed noisy images with no parsing results, yielding 16,253 pairs. The remaining images are further split into a training set and a testing set with 14,221 and 2,032 pairs respectively. Note that during testing, the person should wear a different clothing item than the target one as in real-world scenarios, so we randomly shuffled the clothing product images in these 2,032 test pairs for evaluation.

---
[2]Note that we focus on tops since they are representative in attire with diverse visual graphics and significant deformations. Our method is general and can also be trained for pants, skirts, outerwears, *etc*.

### 4.2. Implementation Details

**Training setup**. Following recent work using encoder-decoder structures [21, 36], we use the Adam [24] optimizer with $\beta_1 = 0.5$, $\beta_2 = 0.999$, and a fixed learning rate of 0.0002. We train the encoder-decoder generator for 15K steps and the refinement network for 6K steps both with a batch size of 16. The resolution of the synthetic samples is $256 \times 192$.

**Encoder-decoder generator**. Our network for the coarse stage contains 6 convolutional layers for encoding and decoding, respectively. All encoding layers consist of $4 \times 4$ spatial filters with a stride of 2, and their numbers of filters are 64, 128, 256, 512, 512, 512, respectively. For decoding, similar $4 \times 4$ spatial filters are adopted with a stride of 1/2 for all layers, whose number of channels are 512, 512, 256, 128, 64, 4. The choice of activation functions and batch normalizations are the same as in [20]. Skip connections [38] are added between encoder and decoder to improve the performance. $\lambda_i$ in Eqn. 1 is chosen to scale the loss of each term properly [6].

**Refinement network**. The network is a four-layer fully convolutional model. Each of the first three layers has $3 \times 3 \times 64$ filters followed by Leaky ReLUs and the last layer outputs the composition mask with $1 \times 1$ spatial filters followed by a sigmoid activation function to scale the output to $(0, 1)$. $\lambda_i$ in Eqn. 4 is the same as in Eqn. 1, $\lambda_{warp} = 0.1$ and $\lambda_{TV} = 5e - 6$.

**Runtime**. The runtime of each component in VITON: Human Parsing (159ms), Pose estimation (220ms), Encoder-Decoder (27ms), TPS (180ms), Refinement (20ms). Results other than TPS are obtained on a K40 GPU. We expect further speed up of TPS when implemented in GPU.

### 4.3. Compared Approaches

To validate the effectiveness of our framework, we compare with the following alternative methods.

**GANs with Person Representation (PRGAN)** [32, 51]. Existing methods that leverage GANs conditioned on either poses [32] or body shape information [51] are not directly comparable since they are not designed for the virtual try-on task. To achieve fair comparisons, we enrich the input of [51, 32] to be the same as our model (a 22-channel representation, $p$ + target clothing image $c$) and adopt their GAN structure to synthesize the reference image.

**Conditional Analogy GAN (CAGAN)** [21]. CAGAN formulates the virtual try-on task as an image analogy problem - it treats the original item and the target clothing item together as a condition when training a Cycle-GAN [50]. However, at test time, it also requires the product image of the original clothing in the reference image, which makes it infeasible in practical scenarios. But we compare with this approach for completeness. Note that for fairness, we modify their encoder-decoder generator to have the same structure as ours, so that it can also generate $256 \times 192$ images. Other implementation details are the same as in [21].

**Cascaded Refinement Network (CRN)** [6]. CRN leverages a cascade of refinement modules, and each module takes the output from its previous module and a down-sampled version of the input to generate a high-resolution synthesized image. Without adversarial training, CRN regresses to a target image using a CNN network. To compare with CRN, we feed the same input of our generator to CRN and output a $256 \times 192$ synthesized image.

**Encoder-decoder generator**. We only use the network of our first stage to generate the target virtual try-on effect, without the TPS warping and the refinement network.

**Non-parametric warped synthesis**. Without using the coarse output of our encoder-decoder generator, we estimate the TPS transformation using shape context matching and paste the warped garment on the reference image. A similar baseline is also presented in [51].

The first three state-of-the-art approaches are directly compared with our encoder-decoder generator without explicitly modeling deformations with warping, while the last Non-parametric warped synthesis method is adopted to demonstrate the importance of learning a composition based on the coarse results.

### 4.4. Qualitative Results

Figure 6 presents a visual comparison of different methods. CRN and encoder-decoder create blurry and coarse results without knowing where and how to render the details of target clothing items. Methods with adversarial training produce shaper edges, but also cause undesirable artifacts. Our Non-parametric baseline directly pastes the warped target image to the person regardless of the inconsistencies between the original and target clothing items, which results in unnatural images. In contrast to these methods, VITON accurately and seamlessly generates detailed virtual try-on results, confirming the effectiveness of our framework.

However, there are some artifacts around the neckline in the last row, which results from the fact that our model cannot determine which regions near the neck should be visible (*e.g*., the neck tag should be hided in the final result, see supplementary material for more discussions). In addition, pants, without providing any product images of them, are also generated by our model. This indicates that our model implicitly learns the co-occurrence between different fash-ion items. VITON is also able to keep the original pants if the pants regions are handled in the similar way as face and hair (*i.e.*, extract pants regions and take them as the input to the encoder). More results and analysis are present in the supplementary material.

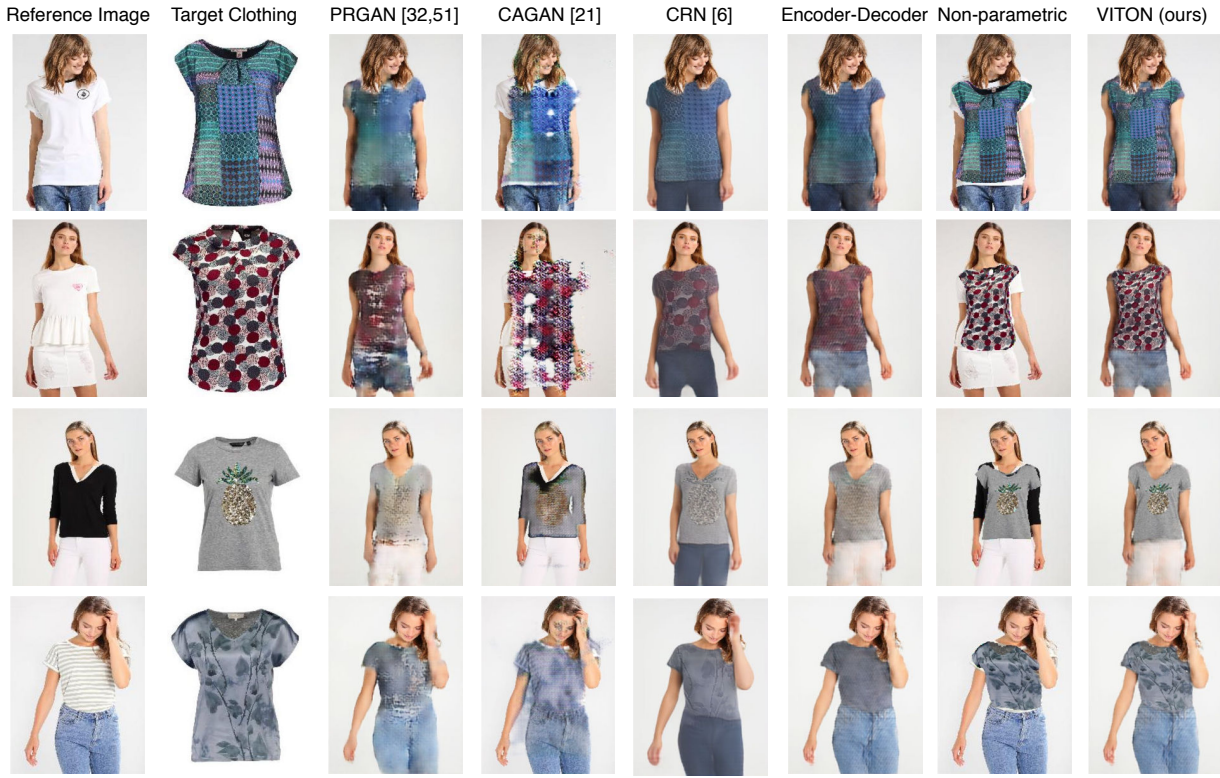

Figure 6: **Qualitative comparisons of different methods**. Our method effectively renders the target clothing on to a person.

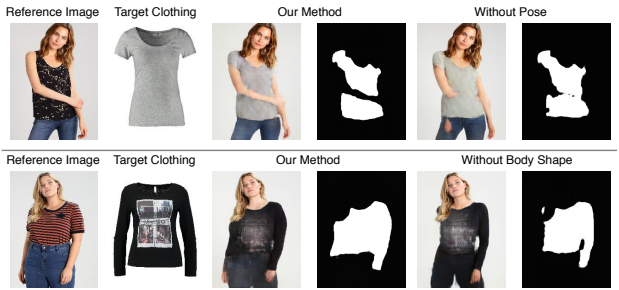

Figure 7: **Effect of removing pose and body shape from the person representation**. For each method, we show its coarse result and predicted clothing mask output by the corresponding encoder-decoder generator.

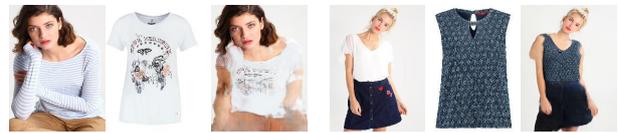

Figure 8: **Failure cases of our method**.

**Person representation analysis**. To investigate the effectiveness of pose and body shape in the person representation, we remove them from the representation individually and compare with our full representation. Sampled coarse results are illustrated in Figure 7. We can see that for a person with a complicated pose, using body shape information alone is not sufficient to handle occlusion and pose ambiguity. Body shape information is also critical to adjust the target item to the right size. This confirms the proposed clothing-agnostic representation is indeed more comprehensive and effective than prior work.

**Failure cases**. Figure 8 demonstrates two failure cases of our method due to rarely-seen poses (example on the left) or a huge mismatch in the current and target clothing shapes (right arm in the right example).

**In the wild results**. In addition to experimenting with constrained images, we also utilize in the wild images from the COCO dataset [29], by cropping human body regions and running our method on them. Sample results are shown in Figure 9, which suggests our method has potentials in applications like generating people in clothing [26].

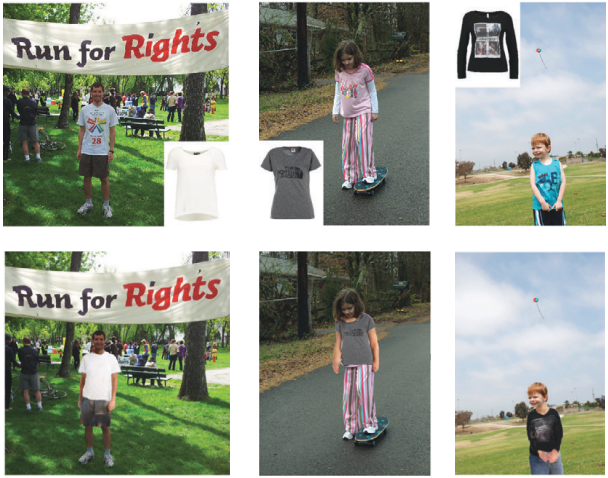

Figure 9: **In the wild results**. Our method is applied to images on COCO.

| Method | IS | Human |
|---|---|---|
| PRGAN [32, 51] | 2.688 ± 0.098 | 27.3% |
| CAGAN [21] | 2.981 ± 0.087 | 21.8% |
| CRN [6] | 2.449 ± 0.070 | 69.1% |
| Encoder-Decoder | 2.455 ± 0.110 | 58.4% |
| Non-parametric | 3.373 ± 0.142 | 46.4% |
| VITON (Ours) | 2.514 ± 0.130 | 77.2% |
| Real Data | 3.312 ± 0.098 | - |

Table 1: Quantitative evaluation on our virtual try-on dataset.

## 4.5. Quantitative Results

We also compare VITON with alternative methods quantitatively based on Inception Score [39] and a user study.

**Inception Score.** Inception Score (IS) [39] is usually used to quantitatively evaluate the synthesis quality of image generation models [32, 33, 47]. Models producing visually diverse and semantically meaningful images will have higher Inception Scores, and this metric correlates well with human evaluations on image datasets like CIFAR10.

**Perceptual user study.** Although Inception Score can be used as an indicator of the image synthesis quality, it cannot reflect whether the details of the target clothing are naturally transferred or the pose and body of the clothed person are preserved in the synthesized image. Thus, simialr to [6, 9], we conducted a user study on the Amazon Mechanical Turk (AMT) platform. On each trial, a worker is given a person image, a target clothing image and two virtual try-on results generated by two different methods (both in $256 \times 192$). The worker is then asked to choose the one that is more realistic and accurate in a virtual try-on situation. Each AMT job contains 5 such trials with a time limit of 200 seconds. The percentage of trials in which one method is rated better than other methods is adopted as the Human evaluation metric following [6] (chance is 50%).

Quantitative comparisons are summarized in Table 1. Note that the human score evaluates whether the virtual try-on results, synthetic images with a person wearing the target item, are realistic. However, we don't have such ground-truth images - the same person in the same pose wearing the target item (IS measures the characteristics of a set, so we use all reference images in the test set to estimate the IS of real data).

According to this table, we make the following observations: (a) Automatic measures like Inception Score are not suitable for evaluating tasks like virtual try-on. The reasons are two-fold. First, these measures tend to reward sharper image content generated by adversarial training or direct image pasting, since they have higher activation values of neurons in Inception model than those of smooth images. This even leads to a higher IS of the Non-parametric baseline over real images. Moreover, they are not aware of the task and cannot measure the desired properties of a virtual try-on system. For example, CRN has the lowest IS, but ranked the 2nd place in the user study. Similar phenomena are also observed in [27, 6]; (b) Person representation guided methods (PRGAN, CRN, Encoder-Decoder, VITON) are preferred by humans. CAGAN and Non-parametric directly take the original person image as inputs, so they cannot deal with cases when there are inconsistencies between the original and target clothing item, *e.g.*, rendering a short-sleeve T-shirt on a person wearing a long-sleeve shirt; (c) By compositing the coarse result with a warped clothing image, VITON performs better than each individual component. VITON also obtains a higher human evaluation score than state-of-the-art generative models and outputs more photo-realistic virtual try-on effects.

To better understand the noise of the study, we follow [6, 32] to perform time-limited (0.25s) real or fake test on AMT, which shows 17.18% generated images are rated as real, and 11.46% real images are rated as generated.

## 5. Conclusion

We presented a virtual try-on network (VITON), which is able to transfer a clothing item in a product image to a person relying only on RGB images. A coarse sample is first generated with a multi-task encoder-decoder conditioned on a detailed clothing-agnostic person representation. The coarse results are further enhanced with a refinement network that learns the optimal composition. We conducted experiments on a newly collected dataset, and promising results are achieved both quantitatively and qualitatively. This indicates that our 2D image-based synthesis pipeline can be used as an alternative to expensive 3D based methods.

# References

[1] U.S. fashion and accessories e-retail revenue 2016-2022. https://www.statista.com/statistics/278890/us-apparel-and-accessories-retail-e-commerce-revenue/. 1

[2] Z. Al-Halah, R. Stiefelhagen, and K. Grauman. Fashion forward: Forecasting visual style in fashion. In *ICCV*, 2017. 2

[3] S. Belongie, J. Malik, and J. Puzicha. Shape matching and object recognition using shape contexts. *IEEE TPAMI*, 2002. 4

[4] F. Bogo, A. Kanazawa, C. Lassner, P. Gehler, J. Romero, and M. J. Black. Keep it smpl: Automatic estimation of 3d human pose and shape from a single image. In *ECCV*, 2016. 1

[5] Z. Cao, T. Simon, S.-E. Wei, and Y. Sheikh. Realtime multi-person 2d pose estimation using part affinity fields. In *CVPR*, 2017. 3

[6] Q. Chen and V. Koltun. Photographic image synthesis with cascaded refinement networks. In *ICCV*, 2017. 2, 6, 8

[7] A. Dosovitskiy and T. Brox. Generating images with perceptual similarity metrics based on deep networks. In *NIPS*, 2016. 4

[8] A. Dosovitskiy, J. Tobias Springenberg, and T. Brox. Learning to generate chairs with convolutional neural networks. In *CVPR*, 2015. 2

[9] C. Gan, Z. Gan, X. He, J. Gao, and L. Deng. Stylenet: Generating attractive visual captions with styles. In *CVPR*, 2017. 8

[10] L. A. Gatys, A. S. Ecker, and M. Bethge. Image style transfer using convolutional neural networks. In *CVPR*, 2016. 5

[11] K. Gong, X. Liang, X. Shen, and L. Lin. Look into person: Self-supervised structure-sensitive learning and a new benchmark for human parsing. In *CVPR*, 2017. 3

[12] I. Goodfellow, J. Pouget-Abadie, M. Mirza, B. Xu, D. Warde-Farley, S. Ozair, A. Courville, and Y. Bengio. Generative adversarial nets. In *NIPS*, 2014. 2

[13] P. Guan, L. Reiss, D. A. Hirshberg, A. Weiss, and M. J. Black. Drape: Dressing any person. *ACM TOG*, 2012. 2

[14] M. Hadi Kiapour, X. Han, S. Lazebnik, A. C. Berg, and T. L. Berg. Where to buy it: Matching street clothing photos in online shops. In *ICCV*, 2015. 2

[15] X. Han, Z. Wu, P. X. Huang, X. Zhang, M. Zhu, Y. Li, Y. Zhao, and L. S. Davis. Automatic spatially-aware fashion concept discovery. In *ICCV*, 2017. 2

[16] X. Han, Z. Wu, Y.-G. Jiang, and L. S. Davis. Learning fashion compatibility with bidirectional lstms. In *ACM Multimedia*, 2017. 2

[17] T. Hassner, S. Harel, E. Paz, and R. Enbar. Effective face frontalization in unconstrained images. In *CVPR*, 2015. 5

[18] A. Hilsmann and P. Eisert. Tracking and retexturing cloth for real-time virtual clothing applications. In *MIRAGE*, 2009. 2

[19] Y. Hu, X. Yi, and L. S. Davis. Collaborative fashion recommendation: a functional tensor factorization approach. In *ACM Multimedia*, 2015. 2

[20] P. Isola, J.-Y. Zhu, T. Zhou, and A. A. Efros. Image-to-image translation with conditional adversarial networks. In *CVPR*, 2017. 2, 4, 6

[21] N. Jetchev and U. Bergmann. The conditional analogy gan: Swapping fashion articles on people images. In *ICCVW*, 2017. 2, 5, 6, 8

[22] J. Johnson, A. Alahi, and L. Fei-Fei. Perceptual losses for real-time style transfer and super-resolution. In *ECCV*, 2016. 4

[23] A. Kanazawa, D. W. Jacobs, and M. Chandraker. Warpnet: Weakly supervised matching for single-view reconstruction. In *CVPR*, 2016. 5

[24] D. Kingma and J. Ba. Adam: A method for stochastic optimization. *arXiv preprint arXiv:1412.6980*, 2014. 6

[25] A. Kovashka, D. Parikh, and K. Grauman. Whittlesearch: Image search with relative attribute feedback. In *CVPR*, 2012. 2

[26] C. Lassner, G. Pons-Moll, and P. V. Gehler. A generative model of people in clothing. In *ICCV*, 2017. 2, 7

[27] C. Ledig, L. Theis, F. Huszár, J. Caballero, A. Cunningham, A. Acosta, A. Aitken, A. Tejani, J. Totz, Z. Wang, and W. Shi. Photo-realistic single image super-resolution using a generative adversarial network. In *CVPR*, 2017. 4, 8

[28] X. Liang, L. Lin, W. Yang, P. Luo, J. Huang, and S. Yan. Clothes co-parsing via joint image segmentation and labeling with application to clothing retrieval. *IEEE TMM*, 2016. 2

[29] T.-Y. Lin, M. Maire, S. Belongie, J. Hays, P. Perona, D. Ramanan, P. Dollár, and C. L. Zitnick. Microsoft coco: Common objects in context. In *ECCV*, 2014. 7

[30] S. Liu, Z. Song, G. Liu, C. Xu, H. Lu, and S. Yan. Street-to-shop: Cross-scenario clothing retrieval via parts alignment and auxiliary set. In *CVPR*, 2012. 2

[31] Z. Liu, P. Luo, S. Qiu, X. Wang, and X. Tang. Deepfashion: Powering robust clothes recognition and retrieval with rich annotations. In *CVPR*, 2016. 2

[32] L. Ma, X. Jia, Q. Sun, B. Schiele, T. Tuytelaars, and L. Van Gool. Pose guided person image generation. In *NIPS*, 2017. 2, 4, 6, 8

[33] A. Odena, C. Olah, and J. Shlens. Conditional image synthesis with auxiliary classifier gans. In *ICML*, 2017. 2, 8

[34] G. Perarnau, J. van de Weijer, B. Raducanu, and J. M. Álvarez. Invertible conditional gans for image editing. In *NIPS Workshop*, 2016. 2

[35] G. Pons-Moll, S. Pujades, S. Hu, and M. Black. Clothcap: Seamless 4d clothing capture and retargeting. *ACM TOG*, 2017. 2

[36] A. Radford, L. Metz, and S. Chintala. Unsupervised representation learning with deep convolutional generative adversarial networks. *arXiv preprint arXiv:1511.06434*, 2015. 2, 6

[37] S. Reed, Z. Akata, X. Yan, L. Logeswaran, B. Schiele, and H. Lee. Generative adversarial text to image synthesis. In *ICML*, 2016. 2

[38] O. Ronneberger, P. Fischer, and T. Brox. U-net: Convolutional networks for biomedical image segmentation. In *MICCAI*, 2015. 4, 6

[39] T. Salimans, I. Goodfellow, W. Zaremba, V. Cheung, A. Radford, and X. Chen. Improved techniques for training gans. In *NIPS*, 2016. 8

[40] M. Sekine, K. Sugita, F. Perbet, B. Stenger, and M. Nishiyama. Virtual fitting by single-shot body shape estimation. In *3D Body Scanning Technologies*, 2014. 1, 2

[41] W. Shen and R. Liu. Learning residual images for face attribute manipulation. *CVPR*, 2017. 2

[42] K. Simonyan and A. Zisserman. Very deep convolutional networks for large-scale image recognition. In *ICLR*, 2015. 4

[43] A. Veit, B. Kovacs, S. Bell, J. McAuley, K. Bala, and S. Belongie. Learning visual clothing style with heterogeneous dyadic co-occurrences. In *CVPR*, 2015. 2

[44] K. Yamaguchi, M. Hadi Kiapour, and T. L. Berg. Paper doll parsing: Retrieving similar styles to parse clothing items. In *ICCV*, 2013. 2

[45] S. Yang, T. Ambert, Z. Pan, K. Wang, L. Yu, T. Berg, and M. C. Lin. Detailed garment recovery from a single-view image. In *ICCV*, 2017. 1, 2

[46] D. Yoo, N. Kim, S. Park, A. S. Paek, and I. S. Kweon. Pixel-level domain transfer. In *ECCV*, 2016. 2

[47] H. Zhang, T. Xu, H. Li, S. Zhang, X. Huang, X. Wang, and D. Metaxas. Stackgan: Text to photo-realistic image synthesis with stacked generative adversarial networks. In *ICCV*, 2017. 8

[48] B. Zhao, J. Feng, X. Wu, and S. Yan. Memory-augmented attribute manipulation networks for interactive fashion search. In *CVPR*, 2017. 2

[49] J.-Y. Zhu, P. Krähenbühl, E. Shechtman, and A. A. Efros. Generative visual manipulation on the natural image manifold. In *ECCV*, 2016. 2

[50] J.-Y. Zhu, T. Park, P. Isola, and A. A. Efros. Unpaired image-to-image translation using cycle-consistent adversarial networks. In *ICCV*, 2017. 2, 6

[51] S. Zhu, S. Fidler, R. Urtasun, D. Lin, and C. L. Chen. Be your own prada: Fashion synthesis with structural coherence. In *ICCV*, 2017. 2, 3, 4, 6, 8

[52] X. Zhu, Z. Lei, J. Yan, D. Yi, and S. Z. Li. High-fidelity pose and expression normalization for face recognition in the wild. In *CVPR*, 2015. 5

# VITON: An Image-based Virtual Try-on Network
# Supplemental Material

## Detailed Network Structures

We illustrate the detailed network structure of our encoder-decoder generator in Figure 1, and that of our refinement network in Figure 2.

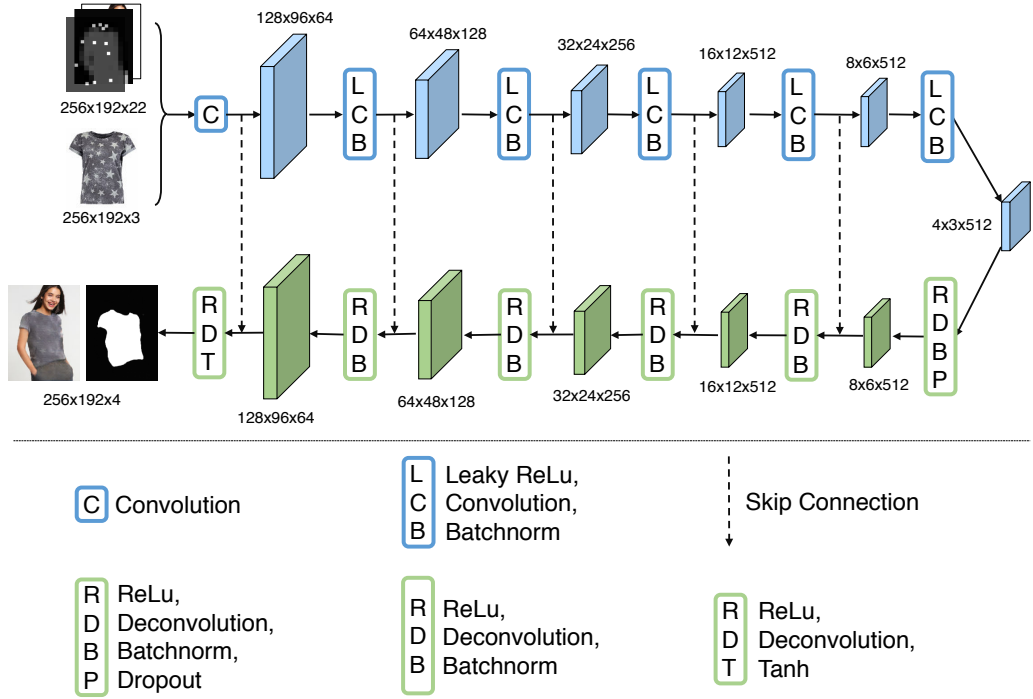

Figure 1: Network structure of our encoder-decoder generator. Blue rectangles indicate the encoding layers and green ones are the decoding layers. *Convolution* denotes $4 \times 4$ convolution with a stride of 2. The negative slope of *Leaky ReLu* is 0.2. *Deconvolution* denotes $4 \times 4$ convolution with a stride of 1/2. The *dropout* probability is set to 0.5.

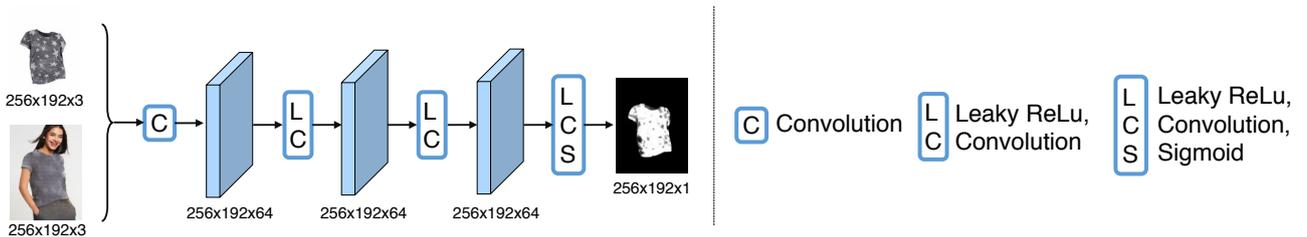

Figure 2: Network structure of our refinement network. *Convolution* denotes $3 \times 3$ convolution with a stride of 1. The negative slope of *Leaky ReLu* is 0.2.

## Person Representation Analysis

To investigate the effectiveness of pose and body shape in the person representation, we remove them individually from our person representation, and train the corresponding encoder-decoder generators to compare with the generator learned by using our full representation (as in Figure 7 in the main paper). Here, we present more qualitative results and analysis.

## Person Representation without Pose

In Figure 3, we show more examples where ignoring the pose representation of the person leads to unsatisfactory virtual try-on results. Although body shape representation and face/hair information are preserved, the model without capturing the person's pose fails to determine where the arms and hands of a person should occur. For example, in the first, third, fifth example in Figure 3, the hands are almost invisible without modeling pose. In the second and forth example, only given the silhouette (body shape) of a person, the model without pose produces ambiguous results when generating the forearms in the virtual try-on results, because it is possible for the forearm to be either in front/back of body or on one side of the body. Note that during these ablation results, we only show the coarse results of each model for simplicity.

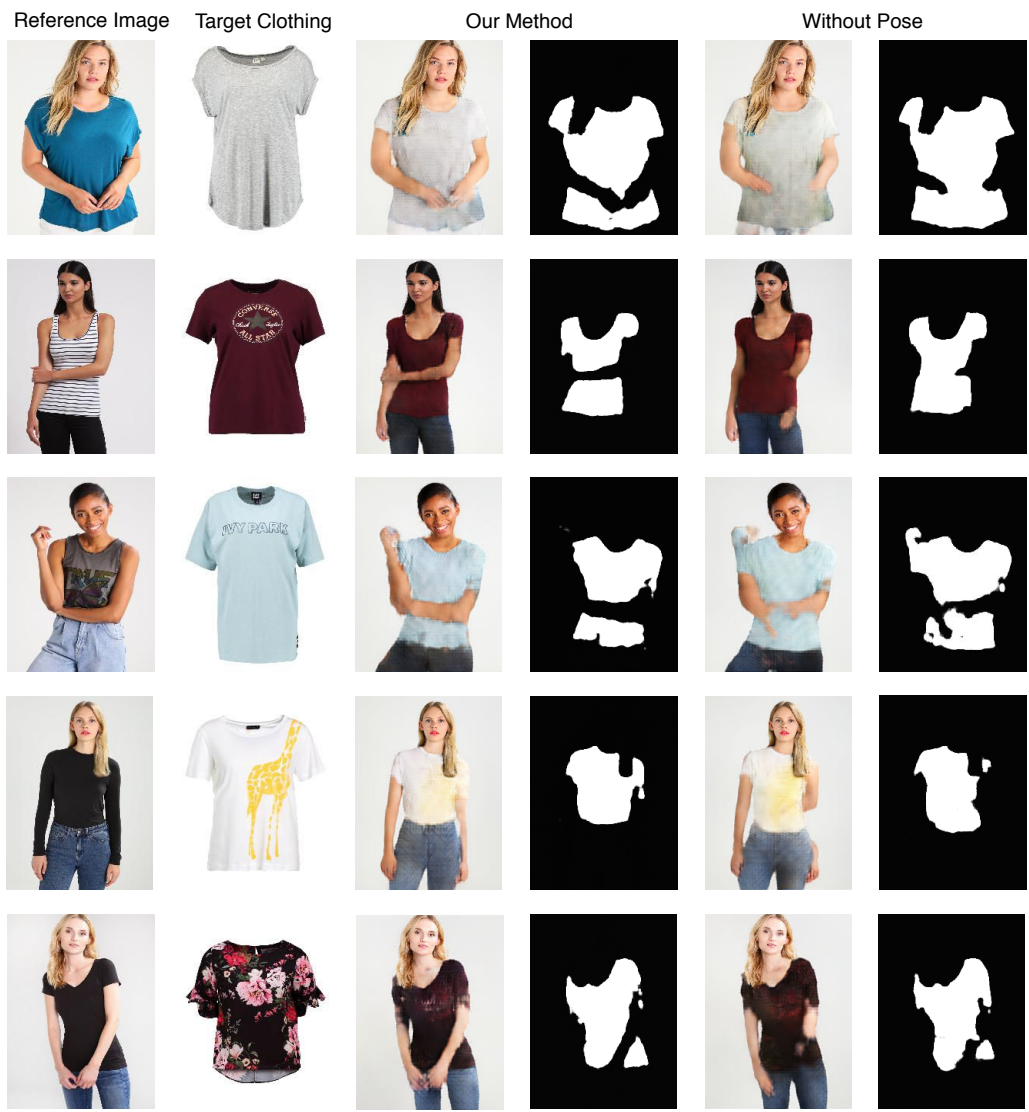

Figure 3: Comparison between the outputs (coarse result + clothing mask) of the encoder-decoder generator trained using our person presentation with the generator trained using the representation without pose.

## Person Representation without Body Shape

Figure 4 demonstrates the cases where body shape is essential to generate a good synthesized result, *e.g.*, virtually trying on plus-size clothing. Interestingly, we also find that body shape can help preserving the poses of the person (second and fifth examples), which suggests that body shape and poses can benefit from each other.

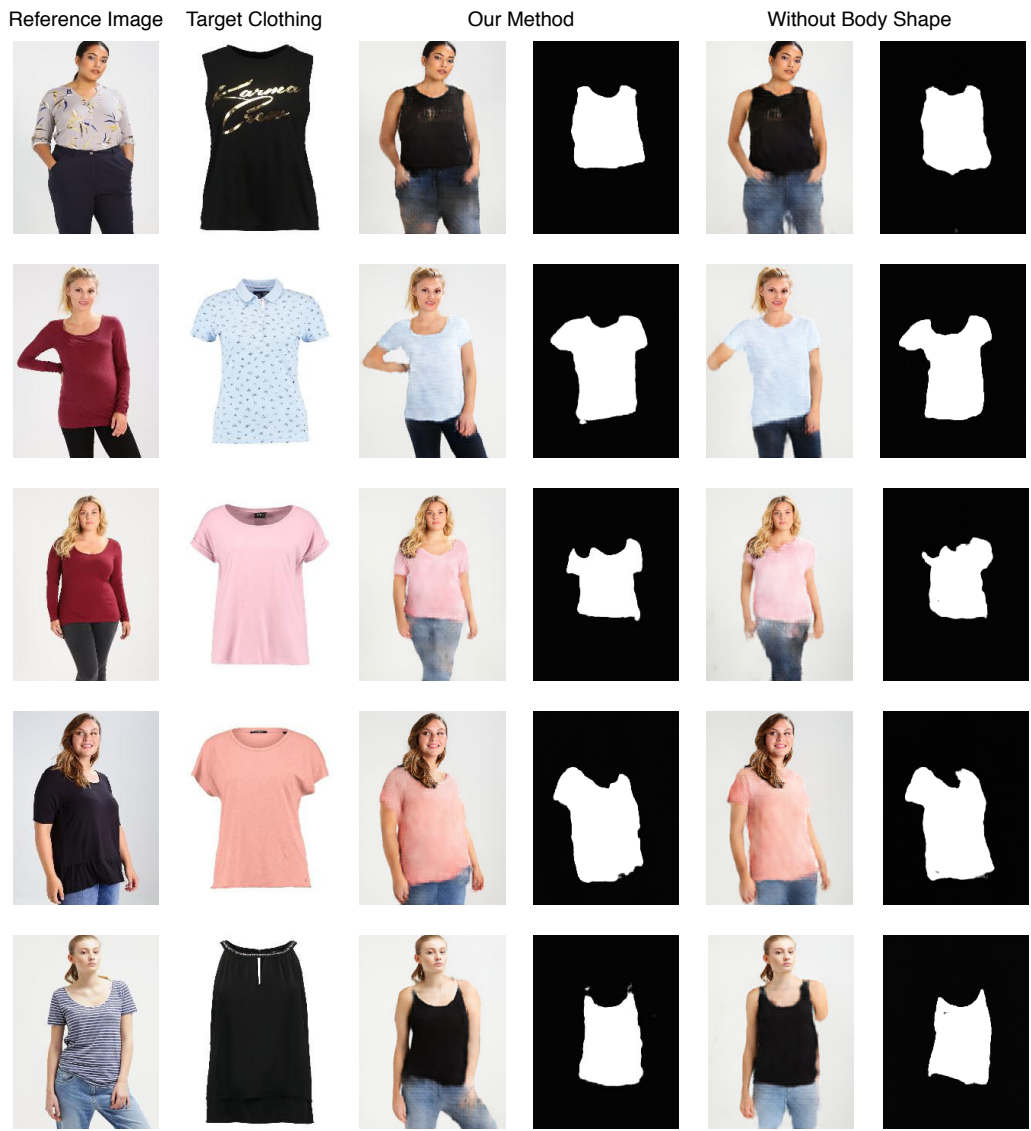

Figure 4: Comparisons between the outputs (coarse result + clothing mask) of the encoder-decoder generator trained using our person presentation with the generator trained using the representation body shape.

## Quantitative evaluation

Furthermore, we conducted the same user study to compare our person representation with the representation without body shape and without pose, respectively. Our method is preferred in 67.6% trials to the representation without body shape, and 77.4% trials to the representation without pose.

## More Qualitative Results

In the following, we present more qualitative comparisons with alternative methods in Figure 5, and more results of our full method in Figure 6. Same conclusions can be drawn as in our main paper.

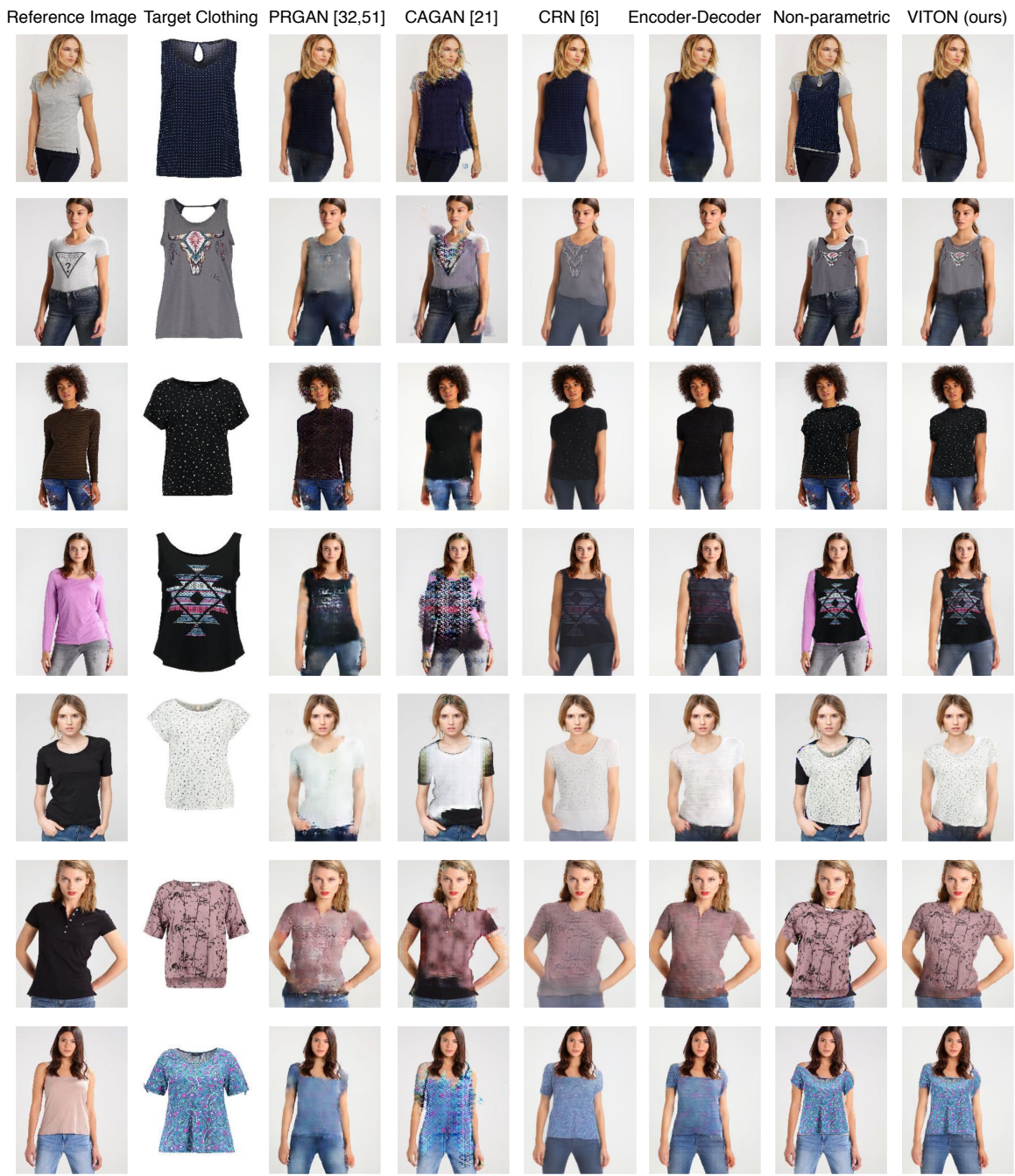

Figure 5: Comparisons of VITON with other methods. Reference numbers correspond to the ones in the main paper.

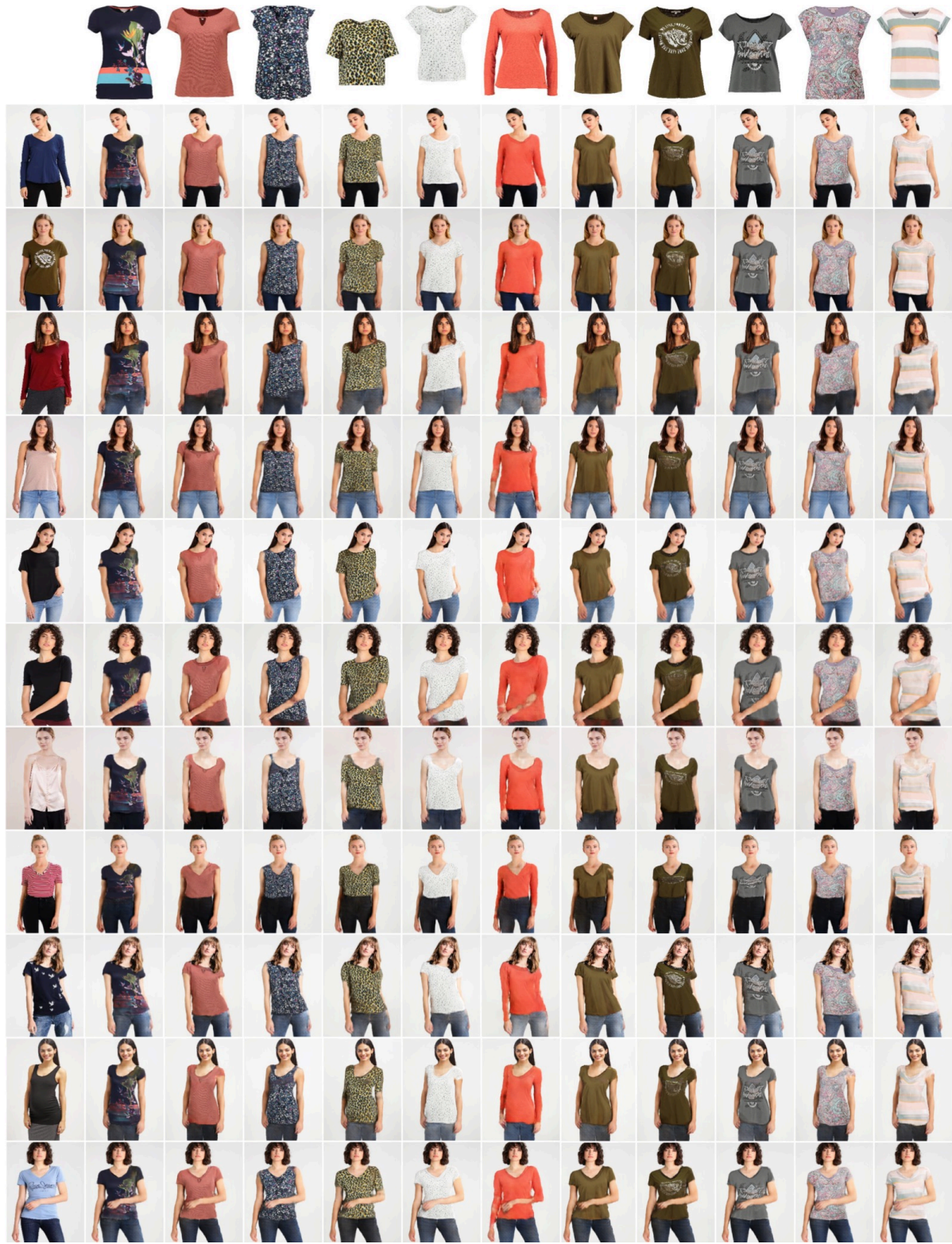

Figure 6: Virtual try-on results of VITON. Each row corresponds to the same person virtually trying on different target clothing items.

## Artifacts near Neck

As mentioned in the main paper, some artifacts may be observed near the neck regions in the synthesized images. This stems from the fact that during warping (See Figure 4 in the main paper) our model warps the whole clothing foreground region as inputs to the refinement network, however regions like neck tag and inner collars should be ignored in the refined result as a result of occlusion.

One way to eliminate these artifacts is to train a segmentation model [2] and remove these regions before warping. We annotated these regions for 2,000 images in our training dataset to train a simple FCN model [2] for this purpose. At test time, this segmentation model is applied to target clothing images to remove the neck regions that should not be visible in the refined results. In this way, we can get rid of the aforementioned artifacts. Some examples are shown in Figure 7. Note that in the main paper, the results are obtained without this segmentation pre-processing step to achieve fair comparisons with alternative methods.

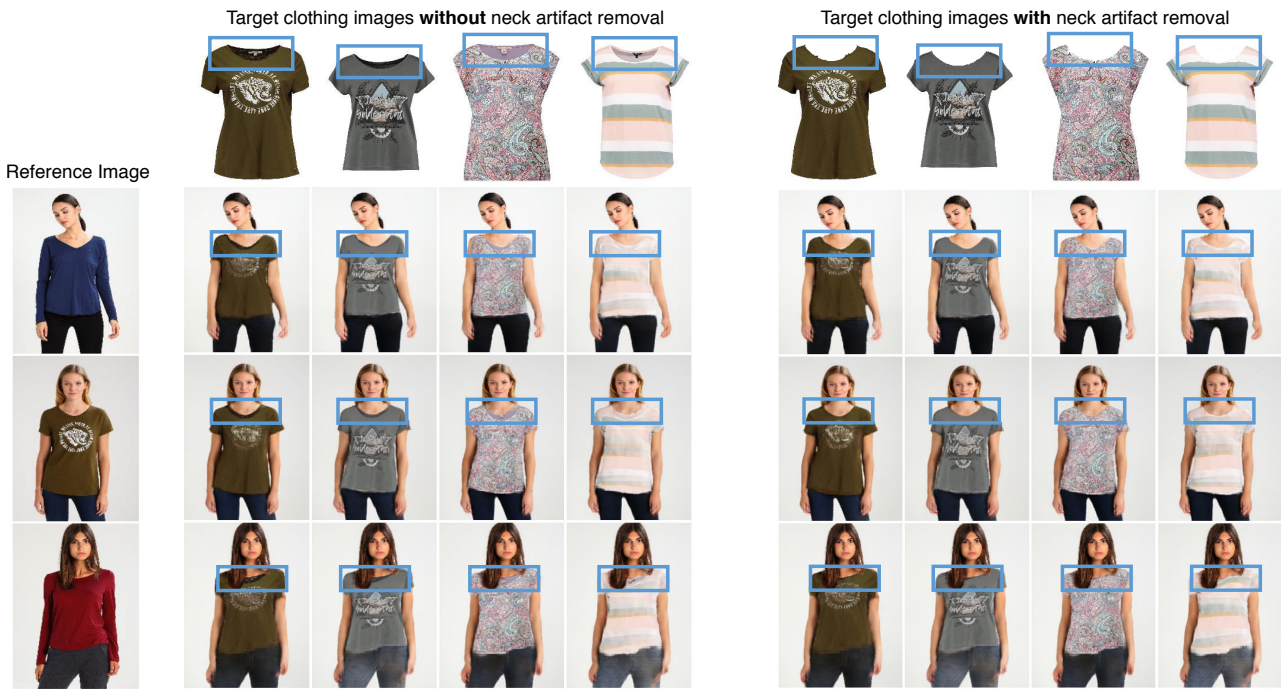

Figure 7: We train an FCN model to segment and remove the regions that cause artifacts during testing. All original results shown in these examples are also included in Figure 6. The artifacts near neck regions (highlighted using blue boxes) are eliminated by removing the unwanted neck areas.

One may notice that there are some discrepancies between the collar style and shape in the predicted image and the target item image. The main reason behind these artifacts is the human parser [1] does not have annotations for necks and treats neck/collar regions as background, and hence the model keeps the original collar style in the reference image. To avoid this, we can augment our human parser with [3] to correctly segment neck regions and retrain our model. Fig 8 illustrates examples of our updated model, which can handle the change in the collars.

## Keep the original pants regions

A straightforward way to keep the original pants regions would be simply keeping the original pixels outside the clothing mask. However, this will cause two problems: (a) if the target item is shorter than the original one, the kept region will be too small, producing a gap between the leg and the target (see the non-parametric results in the 1st and 4th row in Figure 6 of our main paper). Therefore, we re-generate the legs for a seamless overlay. As for the face and hair regions, they are isolated since they are rarely occluded by the clothing; (b) comparisons with baselines like PRGAN and CRN might not be fair, since they generate a whole image without using a mask.

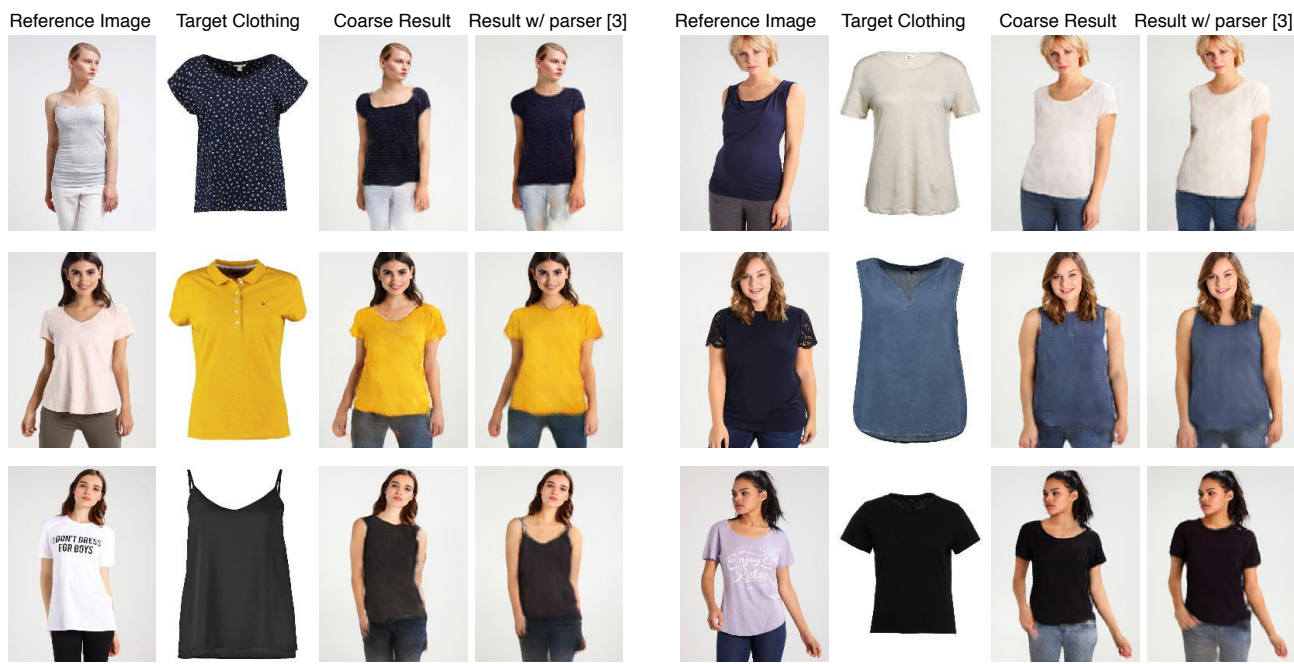

Figure 8: We combine the human parser in [1] and [3] to correct segment neck regions. As a result, the inconsistency between the collar style in the target clothing and the result is properly addressed. Only the coase results are shown for simplicity.

For completeness, we segment the leg region and use it as an input to VITON, the original pants are preserved with some artifacts near the waist regions as shown in Figure 9.

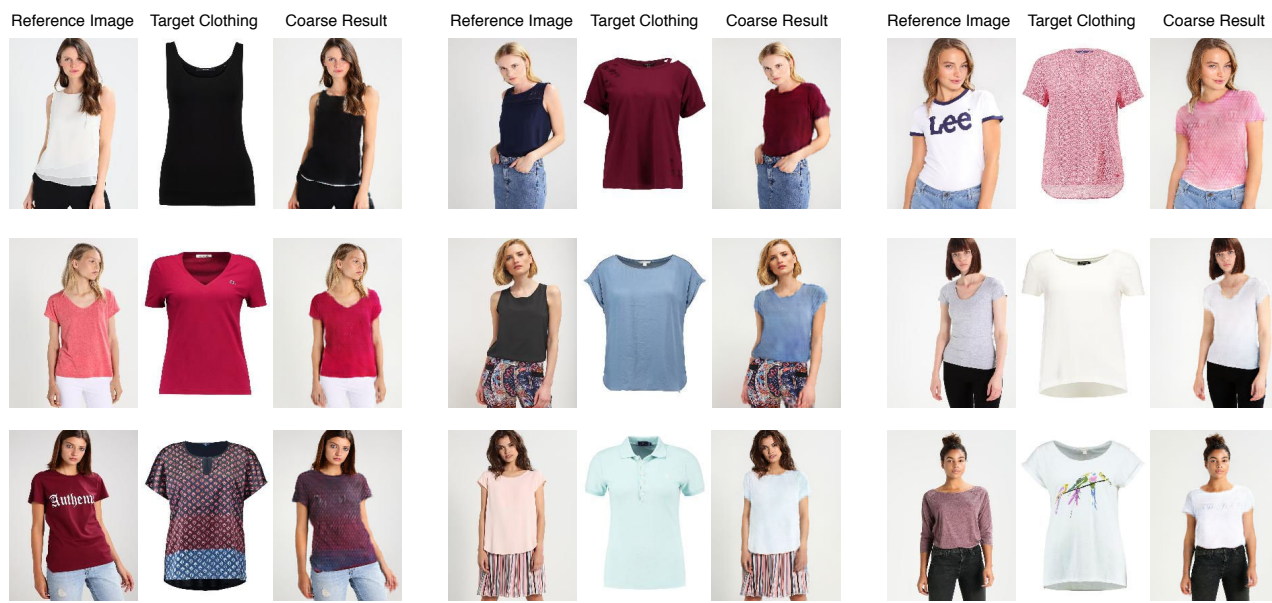

Figure 9: By treating the pants regions in a similar fashion as face/hair, VITON can keep the original pants regions. Only the coase results are shown for simplicity.

## More results

In Figure 10, we show more final (refined) results of VITON with the aforementioned updates - artifacts and inconsistency near neck regions are removed, and original leg regions are kept.

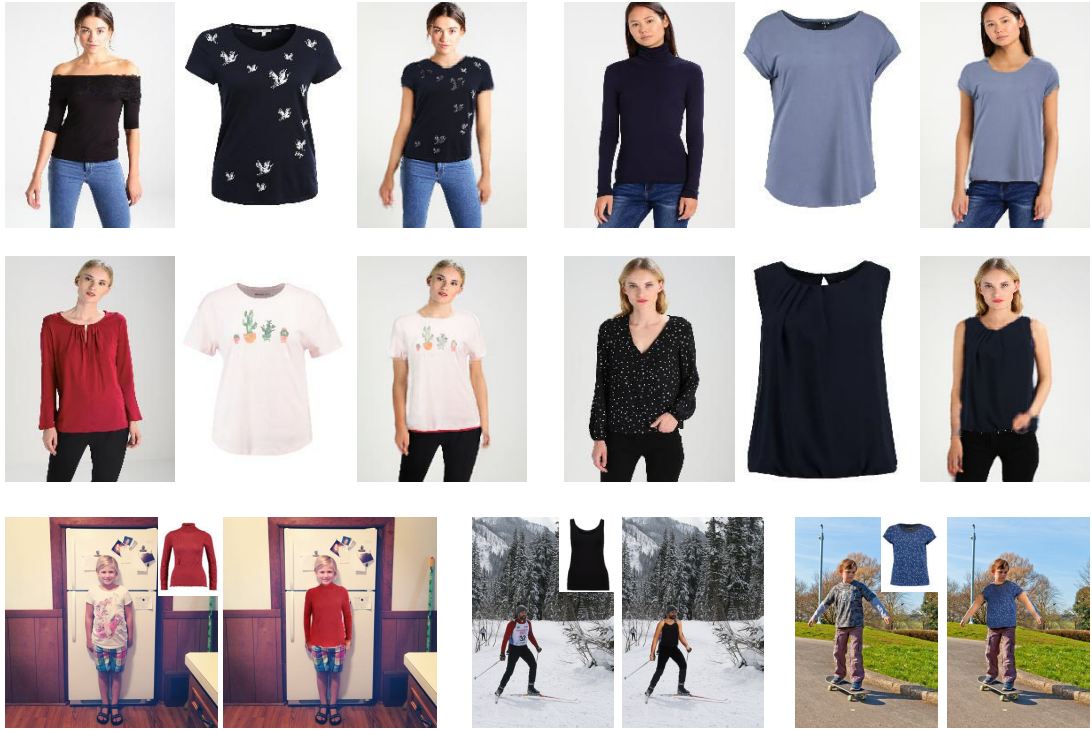

Figure 10: More results of VITON with artifacts near neck region removed and original pants regions preserved. Top two rows are from our visual try-on dataset and the last row contains in-the-wild results from COCO.

# References

[1] K. Gong, X. Liang, X. Shen, and L. Lin. Look into person: Self-supervised structure-sensitive learning and a new benchmark for human parsing. In *CVPR*, 2017. 6, 7

[2] J. Long, E. Shelhamer, and T. Darrell. Fully convolutional networks for semantic segmentation. In *CVPR*, 2015. 6

[3] F. Xia, P. Wang, L.-C. Chen, and A. L. Yuille. Zoom better to see clearer: Human and object parsing with hierarchical auto-zoom net. In *ECCV*, 2016. 6, 7